\def\year{2019}\relax
\documentclass[letterpaper]{article} %DO NOT CHANGE THIS
\usepackage{aaai19}  %Required
\usepackage{times}  %Required
\usepackage{helvet}  %Required
\usepackage{courier}  %Required
\usepackage{url}  %Required
\usepackage{graphicx}  %Required
\usepackage{microtype}
\usepackage{amsmath,amsfonts, amssymb, makecell}
\usepackage{algpseudocode,algorithm,algorithmicx}

\algrenewcommand\algorithmicrequire{\textbf{Input:}}
\algrenewcommand\algorithmicensure{}

\usepackage{subfigure}
\DeclareMathOperator*{\argmax}{argmax}
\DeclareMathOperator*{\minimize}{minimize}

\newtheorem{proposition}{Proposition}
\newcommand\norm[1]{\left\lVert#1\right\rVert}
\begin{document}
\title{Bias Reduction via End-to-End Shift Learning: \\ Application to Citizen Science}
\author{
Di Chen \\
di@cs.cornell.edu\\
Cornell University\\
\And 
Carla P. Gomes\\
gomes@cs.cornell.edu\\
Cornell University\\
%AAAI Press\\
%Association for the Advancement of Artificial Intelligence\\
%2275 East Bayshore Road, Suite 160\\
%Palo Alto, California 94303\\
}
%\interfootnotelinepenalty=10000
\maketitle
\begin{abstract}
Citizen science projects are successful at gathering rich datasets for various applications. 
However, the data collected by citizen scientists are often biased --- in particular, aligned more with the citizens' preferences than with scientific objectives.
We propose the Shift Compensation Network (SCN), an end-to-end learning scheme which learns the shift from the scientific objectives to the biased data while compensating for the shift by re-weighting the training data.
Applied to bird observational data from the citizen science project \textit{eBird}, we demonstrate how SCN quantifies the data distribution shift and outperforms supervised learning models that do not address the data bias.
Compared with competing models in the context of covariate shift, we further demonstrate the advantage of SCN in both its effectiveness and its capability of handling massive high-dimensional data.
\end{abstract}

\section{Introduction}
Citizen science projects \cite{sullivan2014ebird,larrivee2014ebutterfly,seibert2017engaging} play a critical role in collecting rich datasets for scientific research, especially in computational sustainability \cite{gomes2009computational}, because they offer an effective low-cost way to collect large datasets for non-commercial research.
The success of these projects depends heavily on the public's intrinsic motivations as well as the enjoyment of the participants, which engages them to volunteer their efforts \cite{bonney2009citizen}.
Therefore, citizen science projects usually have few restrictions, providing as much freedom as possible to engage volunteers, so that they can decide where, when, and how to collect data, based on their interests. 
As a result, the data collected by volunteers are often biased, 
and align more with their personal preferences, instead of providing systematic observations across various experimental settings.
For example, personal convenience has a significant impact on the data collection process, since the participants contribute their time and effort voluntarily.
Consequently, most data are collected in or near urban areas and along major roads.
On the other hand, most machine learning algorithms are constructed under the assumption that the training data are governed by the same data distribution as that on which the model will later be tested.
As a result, the model trained with biased data would perform poorly when it is evaluated with unbiased test data 
%from a different distribution that is 
designed for the scientific objectives.
%which may hinder which hinders our understanding of
%, which leads to a biased estimation.
\begin{figure}[t]
\centering
\includegraphics[width=7cm]{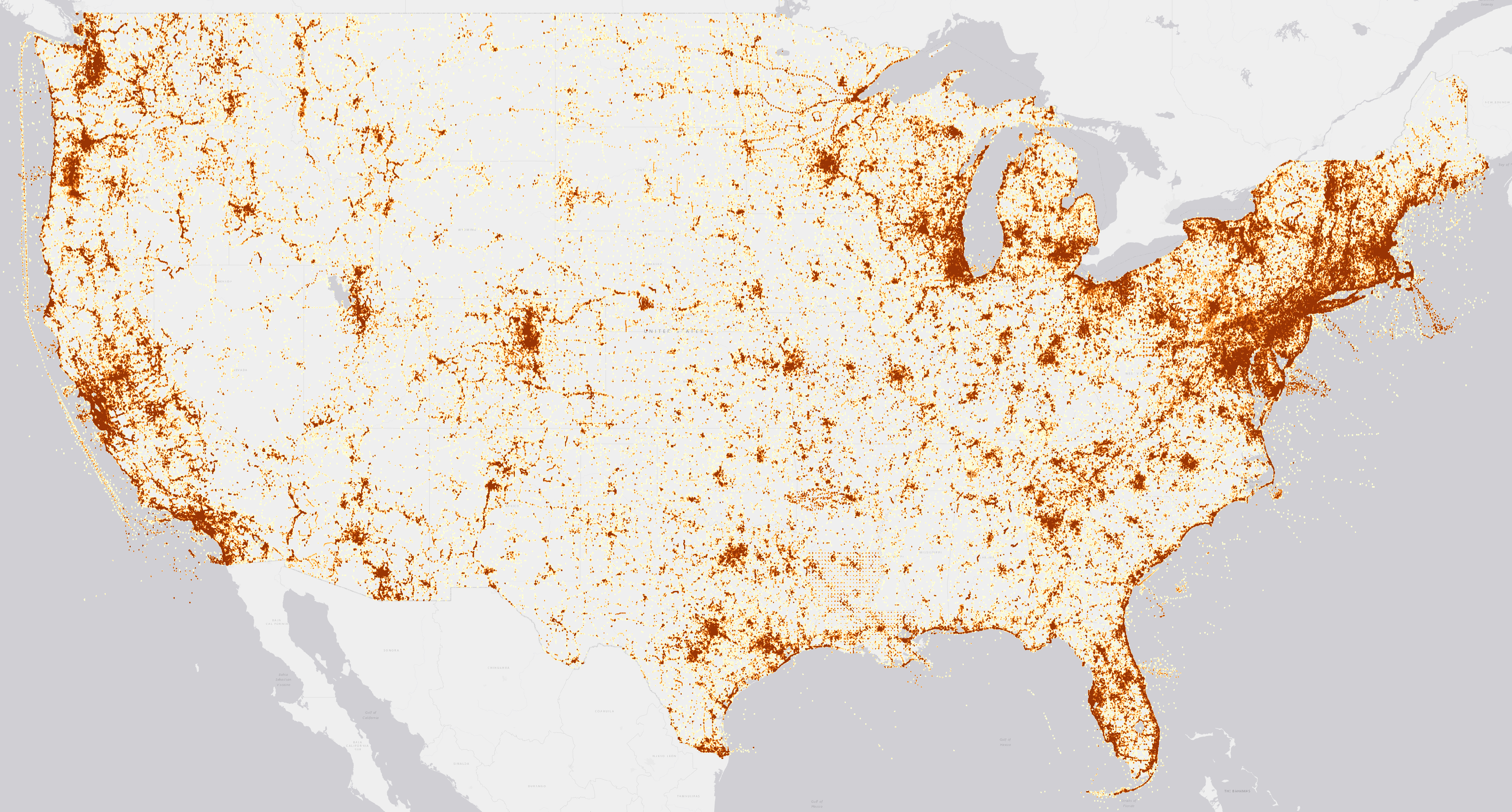}
\caption{Highly biased distribution of \textit{eBird} observations in the continental U.S. Submissions are concentrated in or near urban areas and along major roads.}
\label{fig:map}
\end{figure}

Incentive mechanisms to shift the efforts of volunteers into the more unexplored areas have been proposed \cite{xue2016avicaching}, in order to improve the scientific quality of the data.
However, the scalability of those mechanisms is limited by the budget, and it takes a long time to realize the payback. 
Furthermore, the type of locality also restricts the distribution of collected data. For example, it is difficult to incentivize volunteers to go to remote places, such as deserts or primal forests, to collect data.
Therefore, a tactical learning scheme is needed to bridge the gap between biased data and the desired scientific objectives.

In general, given only the labeled training data (collected by volunteers) and the unlabeled test data (designed for evaluating the scientific objectives), we set out to: 
(i) learn the shift between the training data distribution $P$ (associated with PDF $p(\mathbf{x},y)$) and the test data distribution $Q$ (associated with PDF $q(\mathbf{x},y)$), and
(ii) compensate for that shift so that the model will perform well on the test data. 
To achieve these objectives, we needed to make assumptions on how to bring the training distribution into alignment with the test distribution.
%distribution changes from training to test are necessary. 
Two candidates are \textit{covariate shift} \cite{bickel2009discriminative}, where $p(y|\mathbf{x})=q(y|\mathbf{x})$, and \textit{label shift} \cite {lipton2018detecting}, where $p(\mathbf{x}|y)=q(\mathbf{x}|y)$. 
Motivated by the field observations in the  \textit{eBird} project, where the habitat preference $p(y|\mathbf{x})$ of a given species remains the same throughout a season, while the occurrence records $p(\mathbf{x}|y)$ vary significantly because of the volunteers' preferences, we focused on the \textit{covariate shift} setting.
Informally, covariate shift captures the change in the distribution of the feature (covariate) vector $\mathbf{x}$.
Formally, under covariate shift, we can factor the distributions as follows:
\allowdisplaybreaks
\begin{gather}
    p(\mathbf{x},y) = p(y|\mathbf{x})p(\mathbf{x}) \nonumber\\
    q(\mathbf{x},y) = q(y|\mathbf{x})q(\mathbf{x}) \nonumber\\
    p(y|\mathbf{x})=q(y|\mathbf{x}) \Longrightarrow 
    \frac{q(\mathbf{x},y)}{p(\mathbf{x},y)}=\frac{q(\mathbf{x})}{p(\mathbf{x})}
    \label{eqn:covariate}
\end{gather}
Thus we were able to learn the shift from $P$ to $Q$ and correct our model by quantifying the test-to-training \textbf{shift factor} $q(\mathbf{x})/p(\mathbf{x})$.

\textbf{Our contribution is} \textbf{an end-to-end learning scheme, which we call the Shift Compensation Network (SCN), that estimates the shift factor while re-weighting the training data to correct the model.} 
Specifically, SCN (i) estimates the shift factor by learning a discriminator that distinguishes between the samples drawn from the training distribution and those drawn from the test distribution, and (ii) aligns the mean of the weighted feature space of the training data with the feature space of the test data, which guides the discriminator to improve the quality of the shift factor.
Given the shift factor learned from the discriminator, SCN also compensates for the shift by re-weighting the training samples obtained in the training process to optimize classification loss under the test distribution.
We worked with data from \textit{eBird} \cite{sullivan2014ebird}, which is the world's largest biodiversity-related citizen science project.
Applying SCN to the \textit{eBird} observational data, we demonstrate that it significantly improves multi-species distribution modeling by detecting and correcting for the data bias, thereby providing a better approach for monitoring species distribution as well as for inferring migration changes and global climate fluctuation.
We further demonstrate the advantage of combining the power of discriminative learning and feature space matching, by showing that SCN outperforms all competing models in our experiments.

\section{Preliminaries}
\subsection{Notation}
We use $\mathbf{x} \in \mathcal{X}\subseteq\mathbb{R}^d$ and $y \in \mathcal{Y}=\{0,1,...,l\}$ for the feature and label variables. 
For ease of notation, we use $P$ and $Q$ for the training data and  test data distributions, respectively, defined on $\mathcal{X} \times \mathcal{Y}$.
We use $p(\mathbf{x},y)$ and $q(\mathbf{x},y)$ for the probability density functions (PDFs) associated with $P$ and $Q$, respectively, and $p(\mathbf{x})$ and $q(\mathbf{x})$ for the marginal PDFs of $P$ and $Q$.
\subsection{Problem Setting}
%In this task, we have training data $D_{P}=\{(\mathbf{x}_1, y_1)$, $(\mathbf{x}_2, y_2)$ ...,$(\mathbf{x}_n, y_n)\}$ drawn iid from a training distribution $P$ and test data $D_{Q}=\{\mathbf{x}'_1;\mathbf{x}'_2;...;\mathbf{x}'_n\}$ drawn iid from a test distribution $Q$, where $P$ represents the distribution of the data collected by volunteers and  .
We have labeled training data $D_{P}=\{(\mathbf{x}_1, y_1)$, $(\mathbf{x}_2, y_2)$ ...,$(\mathbf{x}_n, y_n)\}$ drawn iid from a training distribution $P$ and unlabeled test data $D_{Q}=\{\mathbf{x}'_1;\mathbf{x}'_2;...;\mathbf{x}'_n\}$ drawn iid from a test distribution $Q$, where $P$ denotes the data collected by volunteers and $Q$ denotes the data designed for evaluation of the scientific objectives. 
Our goal is to yield good predictions for samples drawn from $Q$.
%To make this task tractable, we have 
Furthermore, we make the following (realistic) assumptions:
%on how the training distribution changes into the test distribution.
\begin{itemize}
    \item $p(y|\mathbf{x}) = q(y|\mathbf{x})$
    \item $p(\mathbf{x})>0$ for every $\mathbf{x} \in \mathcal{X}$ with $q(\mathbf{x})>0$.
\end{itemize}
The first assumption expresses the use of \textit{covariate shift}, which is consistent with the field observations in the \textit{eBird} project. The second assumption ensures that the support of $P$ contains the support of $Q$; without this assumption, this task would not be feasible, as there would be a lack of information on some samples $\mathbf{x}$.

As illustrated in \cite{shimodaira2000improving}, in the covariate shift setting the loss $\ell(f(\mathbf{x}),y)$ on the test distribution $Q$ can be minimized by re-weighting the loss  on the training distribution $P$ with the shift factor $q(\mathbf{x})/p(\mathbf{x})$, that is,
\begin{equation}
    \mathbb{E}_{(\mathbf{x},y)\sim Q}[\ell(f(\mathbf{x}),y)] = \mathbb{E}_{(\mathbf{x},y)\sim P}\Bigg[\ell(f(\mathbf{x}),y)\frac{q(\mathbf{x})}{p(\mathbf{x})}\Bigg]
\end{equation}
Therefore, our goal is to estimate the shift factor $q(\mathbf{x})/p(\mathbf{x})$ and correct the model so that it performs well on $Q$.

\section{End-to-End Shift Learning}
\subsection{Shift Compensation Network}
%One straightforward approach to learn the weights is to directly estimates the distributions $p(x)$ and $q(x)$ from training and test data respectively using kernel density estimation \cite{shimodaira2000improving,sugiyama2005input}.
%However, learning the data distribution $p(x)$ is intrinsically model-based and performs poorly with  high-dimensional data.

\begin{figure}[t]
\centering
\includegraphics[width=7cm]{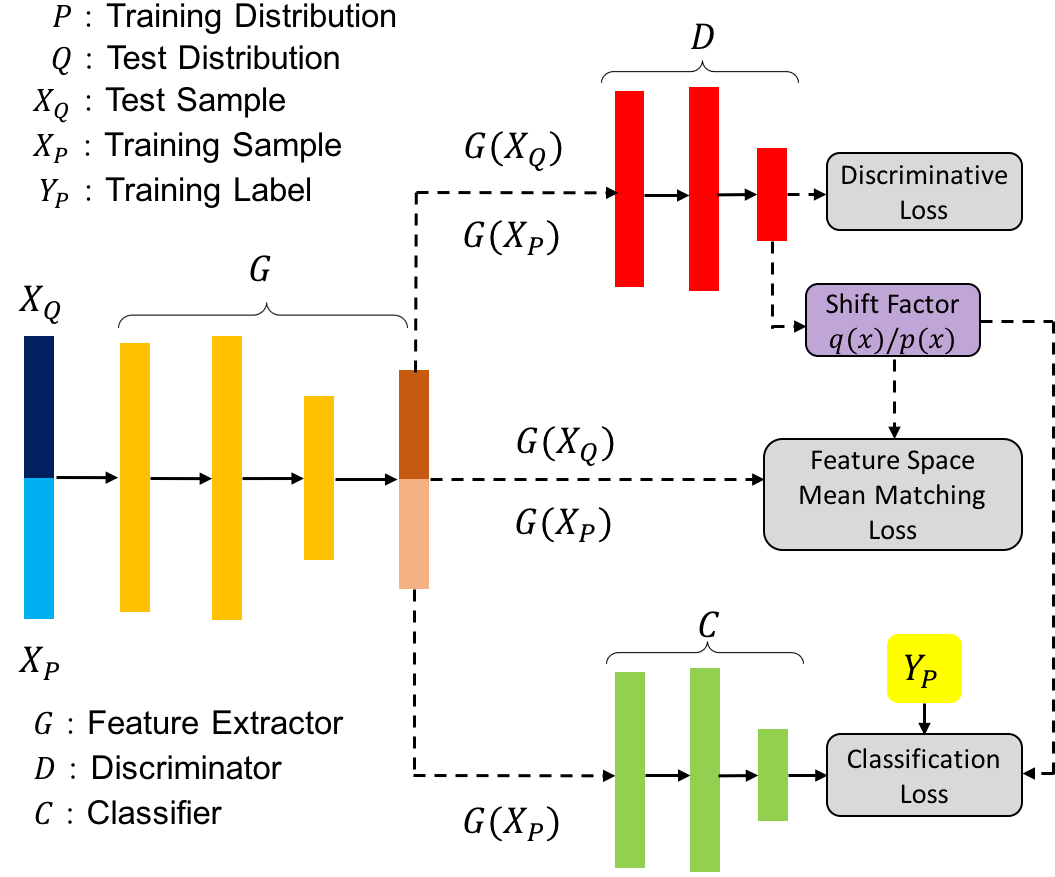}
\caption{Overview of the Shift Compensation Network}
\label{fig:model}
\end{figure}

Fig. \ref{fig:model} depicts the end-to-end learning framework implemented in the Shift Compensation Network (SCN).
A feature extractor $G$ is first applied to encode both the raw training features $X_{P}$ and the raw test features $X_{Q}$ into a high-level feature space. 
Later, we introduce three different losses to estimate the shift factor $q(\mathbf{x})/p(\mathbf{x})$ and optimize the classification task. 

%Inspired by the generative adversarial nets \cite{goodfellow2014generative}, 
We first introduce a discriminative network (with discriminator $D$), together with a discriminative loss, to distinguish between the samples coming from the training data and those coming from the test data.
Specifically, the discriminator $D$ is learned by maximizing the log-likelihood of distinguishing between samples from the training distribution and those from the test distribution, that is, 
\begin{align}
%\resizebox{0.95\hsize}{!}{$
    \mathcal{L}_{D}=&\frac{1}{2}\mathbb{E}_{\mathbf{x}\sim P}[\log(D(G(\mathbf{x})))]+ \\ \nonumber
    &\frac{1}{2}\mathbb{E}_{\mathbf{x}\sim Q}[\log(1-D(G(\mathbf{x})))]
    \label{eqn:lossD}
%$}
\end{align}
\begin{proposition}
\label{prop:one}
For any fixed feature extractor $G$, the optimal discriminator $D^{*}$ for maximizing $\mathcal{L}_{D}$ is
\begin{equation}
    \small D^{*}(G(\mathbf{x}))=\frac{p(\mathbf{x})}{p(\mathbf{x})+q(\mathbf{x})}.
    \nonumber
\end{equation}
Thus we can estimate the shift factor $\frac{q(\mathbf{x})}{p(\mathbf{x})}$ by $\frac{1-D(G(\mathbf{x}))}{D(G(\mathbf{x}))}$.
Proof.
    \begin{align}
%\resizebox{1\hsize}{!}{$
%
    &\resizebox{1\hsize}{!}{$
    D^{*}
     =\argmax\limits_{D}
     \frac{1}{2}\mathbb{E}_{\mathbf{x}\sim P}[\log(D(G(\mathbf{x})))]+
     \frac{1}{2}\mathbb{E}_{\mathbf{x}\sim Q}[\log(1-D(G(\mathbf{x})))]
     $}
    \nonumber \\
     &\resizebox{1\hsize}{!}{$
     \quad\,\,\,=\argmax\limits_{D}\int{p(\mathbf{x})\log(D(G(\mathbf{x})))+q(\mathbf{x})\log(1-D(G(\mathbf{x})))\mbox{d}\mathbf{x}}  
     $}\nonumber\\
     &\small \Longrightarrow \mbox{(maximizing the integrand)}
     \nonumber\\
     &\resizebox{1\hsize}{!}{$
     D^{*}(G(\mathbf{x}))=\argmax\limits_{D(G(\mathbf{x}))}
     p(\mathbf{x})\log(D(G(\mathbf{x})))+q(\mathbf{x})\log(1-D(G(\mathbf{x})))
     $}\nonumber\\
     %
     %&\Longrightarrow \frac{\partial \mathcal{L}_{D} }{\partial D(G(\mathbf{x}))}=0 \nonumber\\
     %&\Longrightarrow \frac{\partial (p(\mathbf{x})\log(D(G(\mathbf{x})))+q(\mathbf{x})\log(1-D(G(\mathbf{x}))) )}{\partial D(G(\mathbf{x}))}=0 \nonumber\\
     &\small \Longrightarrow \mbox{ (the function $d\rightarrow p\log(d)+q\log(1-d)$ achieves its } 
     \nonumber\\
     &\quad \quad \,\, \small \mbox{maximum in $(0,1)$ at $\frac{p}{p+q}$)}
     \nonumber \\
     &\resizebox{0.35\hsize}{!}{$
     D^{*}(G(\mathbf{x}))=\frac{p(\mathbf{x})}{p(\mathbf{x})+q(\mathbf{x})}
     $}
     \nonumber
\end{align}
\label{prop1}
\end{proposition}
%According to the proposition (\ref{prop1})

Our use of a discriminative loss is inspired by the generative adversarial nets (GANs) \cite{goodfellow2014generative}, which have been applied to many areas. 
The fundamental idea of GANs is to train a generator and a discriminator in an adversarial way, where the generator is trying to generate data (e.g., an image) that are as similar to the source data as possible, to fool the discriminator, while the discriminator is trying to distinguish between the generated data and the source data. 
This idea has recently been used in domain adaptation  \cite{tzeng2017adversarial,hoffman2017cycada}, where two generators are trained to align the source data and the target data into a common feature space so that the discriminator cannot distinguish them. 
In contrast to those applications, however, SCN does not have an adversarial training process, where the training and test samples share the same extractor $G$.
In our setting, the training and test distributions share the same feature domain, and they differ only in the frequencies of the sample.
Therefore, instead of trying to fool the discriminator, we want the discriminator to distinguish the training samples from the test samples to the greatest extent possible, so that we can infer the shift factor between the two distributions reversely as in Proposition \ref{prop:one}.

Use of a feature space mean matching (FSMM) loss comes from the notion of kernel mean matching (KMM) \cite{huang2007correcting,gretton2009covariate}, in which the shift factor $w(\mathbf{x})=\frac{q(\mathbf{x})}{p(\mathbf{x})}$ is estimated directly by matching the distributions $P$ and $Q$ in a reproducing kernel Hilbert space (RKHS) $\Phi_{\mathcal{H}}: \mathcal{X}\xrightarrow{} \mathcal{F}$, that is, 
\begin{gather}
    \minimize\limits_{w}{\norm{\mathbb{E}_{\mathbf{x}\sim Q}[\Phi_{\mathcal{H}}(\mathbf{x})]-
    \mathbb{E}_{\mathbf{x}\sim P}[w(\mathbf{x})\Phi_{\mathcal{H}}(\mathbf{x})]
    }_2} \nonumber\\
    \mbox{subject to $w(\mathbf{x})\geq0$ and $\mathbb{E}_{\mathbf{x}\sim P}
[w(\mathbf{x})] = 1$} 
\label{eqn:KMM}
\end{gather}
Though Gretton (\citeyear{gretton2009covariate}) proved that the optimization problem 
(\ref{eqn:KMM}) is convex and has a unique global optimal solution $w(\mathbf{x})=\frac{q(\mathbf{x})}{p(\mathbf{x})}$, the time complexity of KMM is cubic in the size of the training dataset, which is prohibitive when dealing with very large datasets.
We note that even though we do not use the universal kernel \cite{steinwart2002support} in an RKHS, $w(\mathbf{x})=q(\mathbf{x})/p(\mathbf{x})$ still implies that
$$
\norm{\mathbb{E}_{\mathbf{x}\sim Q}[\Phi(\mathbf{x})]-
    \mathbb{E}_{\mathbf{x}\sim P}[w(\mathbf{x})\Phi(\mathbf{x})]
    }_2=0
$$
for any mapping $\Phi(\cdot)$. 
Therefore, our insight is to replace 
%we came up with the idea of replacing 
the $\Phi_{\mathcal{H}(\cdot)}$ with a deep feature extractor $G(\cdot)$ and derive the FSMM loss to further guide the discriminator and improve the quality of $w(\mathbf{x})$.
\begin{gather}
%\resizebox{0.95\hsize}{!}{$ 
    \mathcal{L}_{FSMM}=\norm{\mathbb{E}_{\mathbf{x}\sim Q}[G(\mathbf{x})]-
    \mathbb{E}_{\mathbf{x}\sim P}[w(\mathbf{x})G(\mathbf{x})]
    }_2 \nonumber\\
    =\norm{\mathbb{E}_{\mathbf{x}\sim Q}[G(\mathbf{x})]-
    \mathbb{E}_{\mathbf{x}\sim P}\Bigg[\frac{1-D(G(\mathbf{x}))}{D(G(\mathbf{x}))}G(\mathbf{x})\Bigg]
    }_2 
    \label{eqn:lossFSMM}
%$}
\end{gather}
One advantage of combining the power of $\mathcal{L}_{D}$ and $\mathcal{L}_{FSMM}$ is to prevent overfitting.
Specifically, if we were learning the shift factor $w(\mathbf{x})$ by using only $\mathcal{L}_{FSMM}$, we could end up getting some ill-posed weights $w(\mathbf{x})$ which potentially would not even be relevant to $q(\mathbf{x})/p(\mathbf{x})$. 
This is because 
%One reason for this phenomenon is
the dimensionality of $G(\mathbf{x})$ is usually smaller than the number of training data. 
Therefore, there could be infinitely many solutions of $w(\mathbf{x})$ that achieve zero loss if we consider minimizing $\mathcal{L}_{FSMM}$ as solving a linear equation.
However, with the help of the discriminative loss, which constrains the solution space of equation (\ref{eqn:lossFSMM}), we are able to get good weights which minimize $\mathcal{L}_{FSMM}$ while preventing overfitting.
%, while .
On the other hand, the feature space mean matching loss also plays the role of a regularizer for the discriminative loss, to prevent the discriminator from distinguishing the two distributions by simply memorizing all the samples.
%the feature space mean matching loss, 
Interestingly, in our experiments, we found that the feature space mean matching loss works well empirically even without the discriminative loss.
A detailed comparison will be shown in the Experiments section.

Using the shift factor learned from $\mathcal{L}_{D}$ and $\mathcal{L}_{FSMM}$, we derive the weighted classification loss, that is,
\begin{gather}
    \mathcal{L}_{C}=\mathbb{E}_{\mathbf{x}\sim P}[w(\mathbf{x})\ell(C(G(\mathbf{x})),y)],
    \label{eqn:lossC}
\end{gather}
where $\ell(\cdot,\cdot)$ is typically the cross-entropy loss. 
The classification loss $\mathcal{L}_{C}$ not only is used for the classification task but also ensures that the feature space given by the feature extractor $G$ represents the important characteristics of the raw data.

\subsection{End-to-End Learning for SCN}

\begin{algorithm}[t]
  \caption{Two-step learning in iteration $t$ for SCN 
    \label{alg:SCN}}
  \begin{algorithmic}[1]
    \Require{$X_P$ and $X_Q$ are raw features sampled iid from the training distribution $P$ and the test distribution $Q$; $Y_P$ is the label corresponding to $X_P$;
    $M^{t-1}_P$ and $M^{t-1}_Q$ are the moving averages from the previous iteration.
    }
    \Statex
    \Ensure{\textbf{Step 1}}
    \\
    $m^t_Q = \sum_{\mathbf{x}_i\in X_Q}{G(\mathbf{x}_i)}/|X_Q|$
    \\
    $m^t_P = \sum_{\mathbf{x}_j\in X_P}
    \frac{1-D(G(\mathbf{x}_j))}{D(G(\mathbf{x}_j))}
    G(\mathbf{x}_j)
    /|X_P|$ 
    \\
    $M^{t}_Q \xleftarrow{} \alpha M^{t-1}_Q + (1-\alpha)m^t_Q$
    \\
    $M^{t}_P \xleftarrow{} \alpha M^{t-1}_P + (1-\alpha)m^t_P$
    \\
    $\widehat{M^{t}_Q}\xleftarrow{} M^{t}_Q/(1-\alpha^t)$
    \\
    $\widehat{M^{t}_P}\xleftarrow{} M^{t}_P/(1-\alpha^t)$
    \\
    $\mathcal{L}_{FSMM}\xleftarrow{}\norm{\widehat{M^{t}_Q}-\widehat{M^{t}_P}}_2$
    \\
    $\mathcal{L}_{D}\xleftarrow{} \frac{\sum_{\mathbf{x}_j\in X_P}\log(D(G(\mathbf{x}_j)))}{2|X_P|}+
    \frac{\sum_{\mathbf{x}_i\in X_Q}\log(1-D(G(\mathbf{x}_i)))}{2|X_Q|}$
    \\
    Update the discriminator $D$ and the feature extractor $G$ by ascending along the gradients:
    $$
    \triangledown_{\theta_D}(\lambda_1 \mathcal{L}_{D}-\lambda_2 \mathcal{L}_{FSMM}) \mbox{ and } \triangledown_{\theta_G}(\lambda_1 \mathcal{L}_{D}-\lambda_2 \mathcal{L}_{FSMM})
    $$
    \Statex
    \Ensure{\textbf{Step 2}}
    \\
    For $\mathbf{x}_j\in X_P$, $w(\mathbf{x}_j) \xleftarrow{} \frac{1-D(G(\mathbf{x}_j))}{D(G(\mathbf{x}_j))}$
    \\
    $\mathcal{L}_{C}=\frac{\sum_{\mathbf{x}_j\in X_P}w(\mathbf{x}_j)\ell(C(G(\mathbf{x})),y)}{|X_P|}$
    \\
    Update the classifier $C$ and the feature extractor $G$ by ascending along the gradients:
    $$
    \triangledown_{\theta_C}\mathcal{L}_{C} \mbox{ and } \triangledown_{\theta_G}\mathcal{L}_{C}
    $$
    Here we ignore the gradients coming from the weights $w(\mathbf{x}_j)$, that is, we consider the $w(\mathbf{x}_j)$ as constants.
  \end{algorithmic}
\end{algorithm}

One straightforward way to train SCN is to use mini-batch stochastic gradient descent (SGD) for optimization. 
However, the feature space mean matching loss $\mathcal{L}_{FSMM}$ could have a large variance with small batch sizes. 
For example, if the two sampled batches $X_P$ and $X_Q$ have very few similar features $G(\mathbf{x}_i)$, the $\mathcal{L}_{FSMM}$ could be very large even with the optimal shift factor.
Therefore, instead of estimating $\mathcal{L}_{FSMM}$ based on each mini batch, we maintain the moving averages of both the weighted training features $M_P$ and the test features $M_Q$.
Algorithm \ref{alg:SCN} shows the pseudocode of our two-step learning scheme for SCN.
In the first step, we update the moving averages of both the training data and the test data using the features extracted by $G$ with hyperparameter $\alpha$. 
Then we use the losses $\mathcal{L}_{D}$ and $\mathcal{L}_{FSMM}$ to update the parameters in the feature extractor $G$ and the discriminator $D$ with hyperparameters $\lambda_1$ and $\lambda_2$, respectively, which adjusts the importance of $\mathcal{L}_{D}$ and $\mathcal{L}_{FSMM}$. (We set $\lambda_1=1$ and $\lambda_2=0.1$ in our experiments.)
In the second step, we update the classifier $C$ and the feature extractor $G$ using the estimated shift factor $w(\mathbf{x})$ from the first step.
We initialize the moving averages $M_P$ and $M_Q$ to 0, so that operations (5) and (6) in Algorithm (\ref{alg:SCN}) are applied to compensate for the bias caused by initialization to 0, that is, 
\allowdisplaybreaks
\small
\begin{align}
    \mathbb{E}[M^t_Q]&=\mathbb{E}[\alpha M^{t-1}_Q + (1-\alpha)m^t_Q]
    \nonumber\\
    %&=\mathbb{E}\Bigg[\sum^t_{i=1}(1-\alpha)\alpha^{i-1}m^i_Q\Bigg]
    %\nonumber\\
     &=\sum^t_{i=1}(1-\alpha)\alpha^{i-1}\mathbb{E}[m^i_Q]
    \nonumber\\
    &\approx \widetilde{\mathbb{E}}[m^i_Q](1-\alpha^t)
    \label{eqn:bias}
\end{align}
\normalsize
%The last step in equation (\ref{eqn:bias}) is derived depends on stability of  $\mathbb{E}[m^i_Q]$, which will be achieved at the end of the training. 
Further, since the mini batches are drawn independently, we show that
\small
\allowdisplaybreaks
\begin{align}
    \mbox{Var}[M^t_Q]&=\mbox{Var}[\alpha M^{t-1}_Q + (1-\alpha)m^t_Q]
    \nonumber\\
    %&=\mbox{Var}\Bigg[\sum^t_{i=1}(1-\alpha)\alpha^{i-1}m^i_Q\Bigg]
    %\nonumber\\
     &=\sum^t_{i=1}(1-\alpha)^2\alpha^{2i-2}\mbox{Var}[m^i_Q]
    \nonumber\\
    &\approx \widetilde{\mbox{Var}}[m^i_Q]\frac{1-\alpha}{1+\alpha}(1-\alpha^{2t})
    \label{eqn:var}
\end{align}
\normalsize
That is, by using moving-average estimation, the variance can be reduced by approximately $\frac{1-\alpha}{1+\alpha}$.
Consequently, we can apply an $\alpha$ close to 1 to significantly reduce the variance of $\mathcal{L}_{FSMM}$. 
However, an $\alpha$ close to 1 implies a strong momentum which is too high for the early-stage training. Empirically, we chose $\alpha=0.9$ in our experiments.
%Therefore, we need to balance the tradeoff between smaller variance and stronger momentum.
%Therefore, we could consider increasing the hyperparameter $\alpha$ gradually through the training. 

In the second step of the training process, the shift factor $w(\mathbf{x})$ is used to compensate for the bias between training and test.
Note that we must consider the shift factor as a constant instead of as a function of the discriminator $D$.
Otherwise, minimizing the classification loss  $\mathcal{L}_{C}$ would end up trivially causing all the $w(\mathbf{x})$ to be reduced to zero.

\section{Related Work}
Different approaches for reducing the data bias problem have been proposed.
In mechanism design, 
\citeauthor{xue2016avicaching} (\citeyear{xue2016avicaching}) proposed a two-stage game for providing incentives to shift the efforts of volunteers to more unexplored tasks in order to improve the scientific quality of the data.
In ecology, \citeauthor{phillips2009sample} (\citeyear{phillips2009sample}) improved the modeling of presence-only data by aligning the biases of both training data and background samples. 
\citeauthor{fithian2015bias} (\citeyear{fithian2015bias}) explored the complementary strengths of doing a joint analysis of data coming from different sources to reduce the bias. 
%presence-only and survey
%In domain adaptation, various methods \cite{jiang2007instance,shinnou2015learning,tzeng2017adversarial,hoffman2017cycada} have been proposed to map source and target data to a common feature space while reserving the important characteristics in each domain by reducing the bias between domains.
%which is widely applied to many areas, including language \cite{jiang2007instance,shinnou2015learning}, vision \cite{tzeng2017adversarial}, self-driving \cite{hoffman2017cycada} etc. 
In domain adaptation, various methods \cite{jiang2007instance,shinnou2015learning,tzeng2017adversarial,hoffman2017cycada} have been proposed to reduce the bias between the source domain and the target domain by mapping them to a common feature space while reserving the critical characteristics.

Our work is most closely related to the approaches of  \cite{zadrozny2004learning,huang2007correcting,sugiyama2008direct,gretton2009covariate} developed under the names of \textit{covariate shift} and \textit{sample selection bias}, where the shift factor is learned in order to align the training distribution with the test distribution. 
The earliest work in this domain came from the statistics and econometrics communities, where they addressed the use of non-random samples to estimate behavior.
\citeauthor{heckman1977sample} (\citeyear{heckman1977sample}) addressed sample selection bias, and \citeauthor{manski1977estimation} (\citeyear{manski1977estimation}) investigated estimating parameters under \textit{choice-based bias}, cases that are analogous to a shift in the data distribution.
Later, \cite{shimodaira2000improving} proposed correcting models via weighting of samples in empirical risk minimization (ERM) by the shift factor $q(\mathbf{x})/p(\mathbf{x})$.

One straightforward approach to learning the weights is to directly estimate the distributions $p(\mathbf{x})$ and $q(\mathbf{x})$ from the training and test data respectively, using kernel density estimation \cite{shimodaira2000improving,sugiyama2005input}.
However, learning the data distribution $p(\mathbf{x})$ is intrinsically model based and performs poorly with  high-dimensional data.
\citeauthor{huang2007correcting} (\citeyear{huang2007correcting}) and \citeauthor{gretton2009covariate} (\citeyear{gretton2009covariate})
proposed kernel mean matching (KMM), which estimates the shift factor $w(\mathbf{x}) = q(\mathbf{x})/p(\mathbf{x})$ directly via matching the first moment of the covariate distributions of the training and test data in a reproducing kernel Hilbert space (RKHS) using quadratic programming. 
KLIEP \cite{sugiyama2008direct} estimates $w(\mathbf{x})$ by minimizing the \mbox{Kullback-Leibler} (KL) divergence between the test distribution and the weighted training distribution.
Later, \citeauthor{tsuboi2009direct} (\citeyear{tsuboi2009direct}) derived an extension of KLIEP for applications with a large test set and revealed a close relationship of that approach to kernel mean matching.
Also, \citeauthor{rosenbaum1983central} (\citeyear{rosenbaum1983central}) and \citeauthor{lunceford2004stratification} (\citeyear{lunceford2004stratification}) introduced propensity scoring to design unbiased experiments, which they applied in settings related to sample selection bias.

While the problem of covariate shift has received much attention in the past, it has been used mainly in settings where the size of the dataset is relatively small and the dimensionality of the data is relatively low. 
Therefore, it has not been adequately addressed in settings with massive high-dimensional data, such as hundreds of thousands of high-resolution images.
Among the studies in this area,  \cite{bickel2009discriminative} is the one most closely related to ours.
They tackled this task by modeling the sample selection process using Bayesian inference, where the shift factor is learned by modeling the probability that a sample is selected into training data. 
Though we both use a discriminative model to detect the shift, SCN provides an end-to-end deep learning scheme, where the shift factor and the classification model are learned simultaneously, providing a smoother compensation process, which has considerable advantages for work with massive high-dimensional data and deep learning models.
%which is more suitable and effective for large-scale high-dimensional data, 
In addition, SCN introduces the feature space mean matching loss, which further improves the quality of the shift factor and leads to a better predictive performance.
%as well as the homogeneity between training and test 
%by introducing the feature space mean matching loss, which 
%the mean of their feature spaces.
%What's more, \cite{bickel2009discriminative} tackled this task using Bayesian inference which is not suitable for massive high-dimensional data.
For the sake of fairness, we adapted the work of  \cite{bickel2009discriminative} to the deep learning context in our experiments.

%In a general perspective, our work is also related to the domain adaptation field, which is widely applied to many areas, including language \cite{jiang2007instance,shinnou2015learning}, vision \cite{tzeng2017adversarial}, self-driving \cite{hoffman2017cycada} etc. 

\section{Experiments}

\subsection{Datasets and Implementation Details}
%As one of the most successful citizen science project, e
We worked with a crowd-sourced bird observation dataset from the successful citizen science project \textit{eBird} \cite{sullivan2014ebird}, which is the world's largest biodiversity-related citizen science project, with more than 100 million bird sightings contributed each year by eBirders around the world.
Even though \textit{eBird} amasses large amounts of citizen science data, 
the locations of the collected observation records are highly concentrated in urban areas and along major roads, as shown in Fig. \ref{fig:map}.
This hinders our understanding of species distribution as well as inference of migration changes and global climate fluctuation.
Therefore, we evaluated our SCN
\footnote{Code to reproduce the experiments can be found at {https://bitbucket.org/DiChen9412/aaai2019-scn/}.} 
approach by measuring how we could improve multi-species distribution modeling given biased observational data.
%on the species distribution modeling task using the crowd-sourced bird observation dataset from the successful citizen science project \textit{eBird} \cite{sullivan2014ebird}. 
%\textit{eBird} is the world’s largest biodiversity-related citizen science project, with more than 100 million bird sightings contributed each year by eBirders around the world.

One record in the \textit{eBird} dataset is referred to as a checklist, in which the bird observer records all the species he/she detects as well as the time and the geographical location of the observation site. 
Crossed with the National Land Cover Dataset for the U.S. (NLCD) \cite{homer2015completion}, we obtained a 16-dimensional feature vector for each observation site,  which describes the landscape composition with respect to 16 different land types such as water and forest.
We also collected satellite images for each observation site by matching the geographical location of a site to Google Earth, where several preprocesses have been conducted,  including cloud removal.
Each satellite image covers an area of 17.8 $\mbox{k}\mbox{m}^2$ near the observation site and has 256$\times$256 pixels.
The dataset for our experiments was formed by using all the observation checklists from Bird Conservation Regions (BCRs) 13 and 14 in May from 2002 to 2014, which contains 100,978 observations \cite{us2000north}.
May is a migration period for BCR 13 and 14; therefore a lot of non-native birds pass over this region, which gives us excellent opportunities to observe their choice of habitats during the migration.
We chose the 50 most frequently observed birds as the target species, which cover over 97.4\% of the records in our dataset. 
Because our goal was to learn and predict multi-species distributions across landscapes,
we formed the unbiased test set and the unbiased validation set by overlaying a grid on the map and choosing observation records spatially uniformly.  
We used the rest of the observations to form the spatially biased training set.
Table \ref{table:dataset} presents details of the dataset configuration.

In the experiments, we applied two types of neural networks for the feature extractor $G$: multi-layer fully connected networks (MLPs) and convolutional neural networks (CNNs).
For the NLCD features, we used a three-layer fully connected neural network with hidden units of size 512, 1024 and 512, and with ReLU \cite{nair2010rectified} as the activation function.
For the Google Earth images, we used DenseNet \cite{huang2017densely} with minor adjustments to fit the image size. 
The discriminator $D$ and Classifier $C$ in SCN were all formed by three-layer fully connected networks with hidden units of size 512, 256, and $\#outcome$, and with ReLU as the activation function for the first two layers; there was no activation function for the third layer.
For all models in our experiments, the training process was done for 200 epochs, using a batch size of 128, cross-entropy loss, and an Adam optimizer \cite{kingma2014adam} with a learning rate of 0.0001, and utilized batch normalization \cite{ioffe2015batch}, a 0.8 dropout rate \cite{srivastava2014dropout}, and early stopping to accelerate the training process and prevent overfitting.

%%%%%%%%%%%%%%%%%%%%%%%%%%%%%%%%%%%%%%%%%%%%%%%%%%%%%%%%%%%%%%%
\subsection{Analysis of Performance of the SCN} 

\begin{table}[t]
\newcommand{\tabincell}[2]{\begin{tabular}{@{}#1@{}}#2\end{tabular}}
\centering
\begin{tabular}{|l|l|l|}
\hline
\textbf{Feature Type} & \textbf{NLCD} & \textbf{Google Earth Image}  \\ \hline
\textbf{Dimensionality} & $16$ & $256\times256\times3$  \\ \hline
\textbf{\#Training Set} & 79060 & 79060  \\
\hline
\textbf{\#Validation Set} & 10959 & 10959  \\
\hline
\textbf{\#Test Set} & 10959 & 10959  \\
\hline
\textbf{\#Labels} & 50 & 50   \\
\hline
\end{tabular}
\caption{Statistics of the \textit{eBird} dataset}
\label{table:dataset}
\end{table}

We compared the performance of SCN with baseline models from two different groups.
The first group included \textit{models that ignore the covariate shift} ( which we refer to as vanilla models), that is, models are trained directly by using batches sampled uniformly from the training set without correcting for the bias. 
The second group included \textit{different competitive models for solving the covariate shift problem}: 
(1) kernel density estimation (KDE) methods \cite{shimodaira2000improving,sugiyama2005input}, 
%(2) kernel mean matching (KMM) \cite{huang2007correcting,gretton2009covariate}, 
(2) the \mbox{Kullback-Leibler} Importance Estimation Procedure (KLIEP) \cite{sugiyama2008direct}, and (3) discriminative factor weighting (DFW) \cite{bickel2009discriminative}.
The DFW method was implemented initially by using a Bayesian model, which we adapted to the deep learning model in order to use it with the $eBird$ dataset. 
We did not compare SCN with the kernel mean matching (KMM) methods \cite{huang2007correcting,gretton2009covariate}, because KMM, like many kernel methods, requires the construction and inversion of an $n \times n$ Gram matrix, which has a complexity of $\mathcal{O}(n^3)$. 
This hinders its application to real-life applications, where the value of $n$ will often be in the hundreds of thousands. 
In our experiments, we found that the largest $n$ for which we could feasibly run the KMM code is roughly 10,000 (even with SGD), which is only $10\%$ of our dataset.
To make a fair comparison, we did a grid search for the hyperparameters of all the baseline models to saturate their performance.
Moreover, the structure of the networks for the feature extractor and the classifier used in all the baseline models, were the same as those in our SCN (i.e., $G$ and $C$), while the shift factors for those models were learned using their methods.

\begin{table}[t]
\footnotesize
\centering
\begin{tabular}{|c|c|c|c|}
\hline
 \multicolumn{4}{|c|}{\textbf{NLCD Feature}}  \\
\hline
\textbf{Test Metrics (\%)} & \textbf{AUC} & \textbf{AP} & \textbf{F1 score} \\ 
\hline
vanilla model &  77.86 & 63.31  & 54.90  \\
\hline
SCN & \textbf{80.34} & \textbf{66.17}  & \textbf{57.06}  \\
\hline
KLIEP & 78.87 & 64.33  & 55.63  \\
\hline
KDE & 78.96 & 64.42  & 55.27  \\
\hline
DFW & 79.38 & 64.98 & 55.79  \\
\hline

 \multicolumn{4}{|c|}{\textbf{Google Earth Image}}  \\
\hline
vanilla model &  80.93 & 67.33  & 59.97  \\
\hline
SCN & \textbf{83.80} & \textbf{70.39}  & \textbf{62.37}  \\
\hline
KLIEP & 81.17 & 67.86  & 60.23  \\
\hline
KDE & 80.95 & 67.42  & 60.01  \\
\hline
DFW & 81.99 & 68.44 & 60.77  \\
\hline
\end{tabular}
\caption{Comparison of predictive performance of different methods under three different metrics. (The larger, the better.)
}
\label{table:results}
\end{table}

Table \ref{table:results} shows the average performance of SCN and other baseline models with respect to three different metrics: (1) \textbf{AUC}, area under the ROC curve; (2) \textbf{AP}, area under the precision--recall curve; (3) \textbf{F1 score}, the harmonic mean of precision and recall.
Because our task is a multi-label classification problem, these three metrics were averaged over all 50 species in the datasets.
In our experiments, the standard error of all the models was less than $0.2\%$ under all three metrics.
There are two key results in Table \ref{table:results}: 
\textbf{(1)  All bias-correction models outperformed the vanilla models under all metrics, which shows a significant advantage of correcting for the covariate shift.}
\textbf{(2)  SCN outperformed all the other bias-correcting models, especially on high-dimensional Google Earth images.}

\begin{figure}[t]
\centering
\includegraphics[width=8cm]{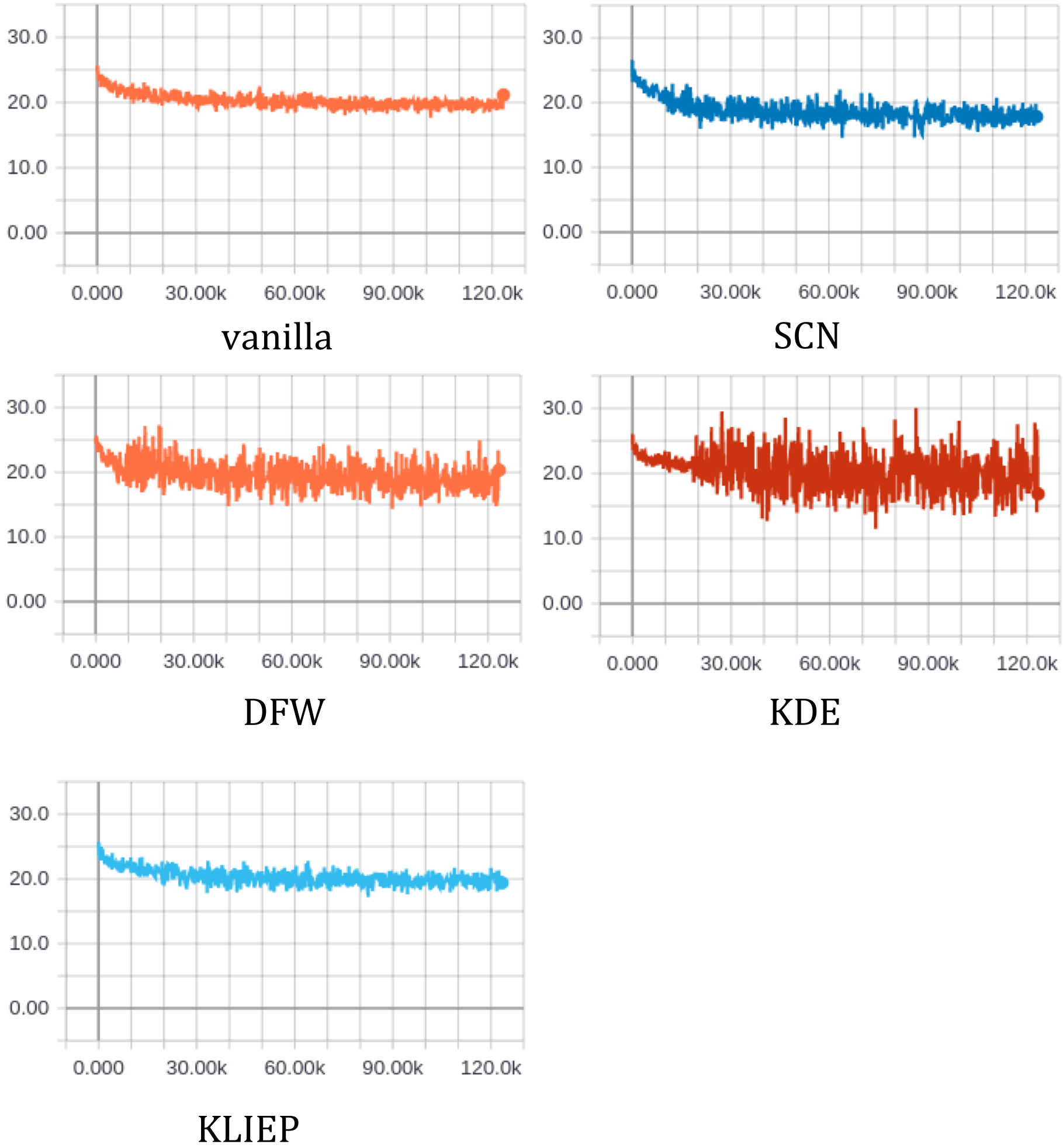}
\caption{The learning curves of all models. The vertical axis shows the cross-entropy loss, and the horizontal axis shows the number of iterations. }
\label{fig:curve}
\end{figure}
The kernel density estimation (KDE) models had the worst performance, especially on Google Earth images.
%, which is consistent with the conclusion in \cite{bickel2009discriminative}.
This is not only because of the difficulty of modeling high-dimensional data distributions, but also because of the sensitivity of the KDE approach. When $p(\mathbf{x}) \ll q(\mathbf{x})$, a tiny perturbation of $p(\mathbf{x})$ could result in a huge fluctuation in the shift factor $q(\mathbf{x})/p(\mathbf{x})$.
KLIEP performed slightly better than KDE, by learning the shift factor $w(\mathbf{x})=\frac{q(\mathbf{x})}{p(\mathbf{x})}$ directly, where it minimized the KL divergence between the weighted training distribution and the test distribution. 
However, it showed only a slight improvement over the vanilla models on Google Earth images.
%However, given limited training and test data, we can only estimate the KL-divergence using an empirical approximation.
%Among 
%One reason that the classical methods KDE and KLIEP performs worse than SCN and DFW is that these two methods focus on finding the best 
DFW performed better than the other two baseline models, which is not surprising, given that DFW learns the shift factor by using a discriminator similar to the one in SCN.
SCN outperformed DFW not only because it uses an additional loss, the feature space mean matching loss, but also because of its end-to-end training process. 
DFW first learns the shift factor by optimizing the discriminator, and then it trains the classification model using samples weighted by the shift factor. 
However, SCN learns the classifier $C$ and the discriminator $D$ simultaneously, where the weighted training distribution approaches the test distribution smoothly through the training process, which performed better empirically than directly adding the optimized weights to the training samples. 
\citeauthor{wang2018cost} (\citeyear{wang2018cost}) also discovered a similar phenomenon in cost-sensitive learning, where pre-training of the neural network with unweighted samples significantly improved the model performance.
One possible explanation of this phenomenon is that the training of deep neural networks depends highly on mini-batch SGD, so that the fluctuation of gradients caused by the weights may hinder the stability of the training process, especially during the early stage of the training.
%as well as the final performance. 
Fig. \ref{fig:curve} shows the learning curves of all five models, where we used the same batch size and an Adam optimizer with the same learning rate.
%and pre-train the model with unweighted samples for 20 epochs as the initialization.
As seen there, SCN had a milder oscillation curve than DFW, which is consistent with the conjecture we stated earlier. 
In our experiments, we pre-trained the baseline models with unweighted samples for 20 epochs in order to achieve a stable initialization. 
Otherwise, some of them would end up falling into a bad local minimum, where they would perform even worse than the vanilla models.

%\subsection{Shift Factor Analysis} 
%

To further explore the functionality of the discriminative loss and the feature mean matching loss in SCN, we implemented several variants of the original SCN model:
\begin{itemize}
    \item SCN: The original Shift Compensation Network
    \item SCN\_D: The Shift Compensation Network without the feature space mean matching loss ($\lambda_2=0$)
    \item SCN\_FSMM: The Shift Compensation Network without the discriminative loss ($\lambda_1=0$)
    \item SCN$^{-}$: The Shift Compensation Network without using moving-average estimation for the feature space mean matching loss ($\alpha=0$)
\end{itemize}
\begin{table}[t]
\footnotesize
\centering
\begin{tabular}{|c|c|c|c|}
\hline
 \multicolumn{4}{|c|}{\textbf{NLCD Feature}}  \\
\hline
\textbf{Test Metrics (\%)} & \textbf{AUC} & \textbf{AP} & \textbf{F1-score} \\ 
\hline
SCN & \textbf{80.34} & \textbf{66.17}  & \textbf{57.06}  \\
\hline
SCN\_D & 79.53 & 65.11  & 56.11  \\
\hline
SCN\_FSMM & 79.58 & 65.17  & 56.26  \\
\hline
SCN$^{-}$ & 80.09 & 65.97 & 56.83  \\
\hline
\multicolumn{4}{|c|}{\textbf{Google Earth Image}}  \\
\hline
SCN & \textbf{83.80} & \textbf{70.39}  & \textbf{62.37}  \\
\hline
SCN\_D & 82.35 & 68.96  & 61.23  \\
\hline
SCN\_FSMM & 82.49 & 69.05  & 61.51  \\
\hline
SCN$^{-}$ & 83.44 & 69.72 & 62.01  \\
\hline
\end{tabular}
\caption{Comparison of predictive performance of the different variants of SCN
}
\label{table:variants}
\end{table}
Table \ref{table:variants} compares the performance of the different variants of SCN, where we observe the following:
(1) Both the discriminative loss and the feature space mean matching loss play an important role in learning the shift factor. 
(2) %By comparing the performance between SCN and SCN\_D+FSMM*,
The moving-average estimation for the feature space mean matching loss shows an advantage over the batch-wise estimation (compare SCN to SCN$^{-}$).
(3) Crossed with Table \ref{table:results}, SCN performs better than DFW, even with only the discriminative loss, which shows the benefit of fitting the shift factor gradually through the training process.
(4) Surprisingly, even if we use only the feature space mean matching loss, which would potentially lead to ill-posed weights, SCN\_FSMM still shows much better performance than the other baselines.
    %the 
    %advantage of learning shift factors and the classifier simultaneously, where 
%shows the performance of those variants. 
%One can see, both the discriminative loss and the feature space mean matching loss play an important role in learning the shift factor. 
%By comparing the performance between SCN and SCN\_D+FSMM*, we can confirm that using moving average to estimate the feature space mean matching loss performs better than using batch-wise estimation. 

\subsection{Shift Factor Analysis} 
\begin{figure}[t]
\centering
\includegraphics[width=8cm]{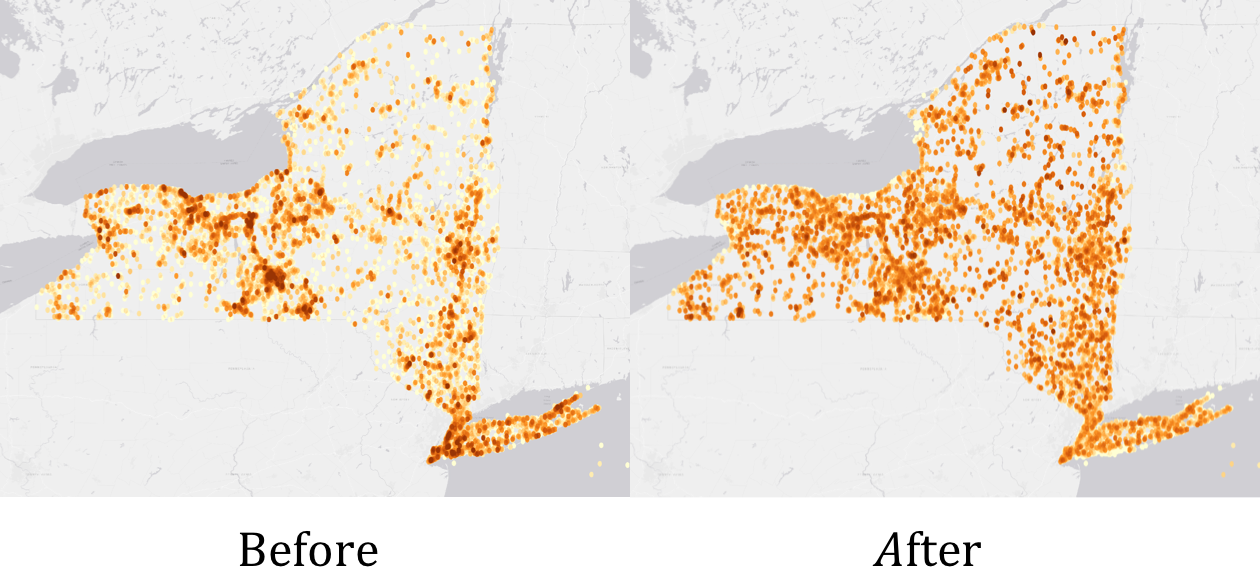}
\caption{Heatmap of the observation records for the month of May in New York State, where the left panel shows the distribution of the original samples and the right one shows the distribution weighted with the shift factor}
\label{fig:factor}
\end{figure}

We visualized the heatmap
%\footnote{The basemap we used for generating the heatmap is from {http://felix.rohrba.ch/en/2016/awesome-basemap-layer-for-your-qgis-project/}} 
of the observation records for the month of May in New York State (Fig. \ref{fig:factor}), where the left panel shows the distribution of the original samples and the right one shows the distribution weighted with the shift factor.
The colors from white to brown indicates the sample popularity from low to high using a logarithmic scale from 1 to 256.
%
%shift factor (Fig.(\ref{fig:factor})) learned from SCN by plotting the output of the discriminator, i.e., $D(G(\mathbf{x}))\approx \frac{p(\mathbf{x})}{p(\mathbf{x})+q(\mathbf{x})}=\frac{1}{w(\mathbf{x})+1}$ in New York state, which is positive correlated to the reciprocal of the shift factor while having a bounded value between $[0,1]$. 
%Informally, this represents the understanding of SCN about how popular the place is compared with the average.
%understanding of SCN about how  
As seen there, the original samples are concentrated in the southeastern portion and Long Island, while the weighted one is more balanced over the whole state after applying the shift factor. 
This illustrates that SCN learns the shift correctly and provides a more balanced sample distribution by compensating for the shift.
%that the understanding of SCN is consistent with the field observation, compared with the heatmap of the observation records.

We investigated the shift factors learned from the different models (Table \ref{table:dis}) by analyzing the ratio of the feature space mean matching loss to the dimensionality of the feature space using equation (\ref{eqn:dis}). 
%One can see, the most observations are mainly swarmed around the big city, especially, the New York city, which is consistent with the field observation.
\begin{gather}
    \norm{\mathbb{E}_{\mathbf{x}\sim Q}[\Phi(\mathbf{x})]-
    \mathbb{E}_{\mathbf{x}\sim P}[w(\mathbf{x})\Phi(\mathbf{x})]
    }_2/\mbox{dim}(\Phi(\mathbf{x})) \nonumber\\
    \resizebox{0.95\hsize}{!}{$ 
    \approx\norm{\frac{1}{|X_Q|}\sum_{i}\Phi(\mathbf{x}_i)-
    \frac{1}{|X_P|}\sum_{i}w(\mathbf{x}_i)\Phi(\mathbf{x}_i)
    }_2/\mbox{dim}(\Phi(\mathbf{x})) 
    $}\label{eqn:dis}
\end{gather}
Here, we chose the output of the feature extractor in each model (such as $G(\mathbf{x})$ in SCN) as the feature space $\Phi(\mathbf{x})$. 
Compared to the vanilla models, all the shift correction models significantly reduced the discrepancy between the weighted training distribution and the test distribution.
However, crossed with Table \ref{table:results}, it is interesting to see the counter-intuitive result that the models with the smaller feature space discrepancies (KDE \& KLIEP) did not necessarily perform better. 
\begin{table}[t]
\footnotesize
\centering
\begin{tabular}{|c|c|}
\hline
 \multicolumn{2}{|c|}{\textbf{Averaged Feature Space Mean Matching Loss}}  \\
%\hline
%\textbf{} & \textbf{L2} & \textbf{L1}  \\ 
\hline
vanilla model &  0.8006   \\
\hline
SCN & 0.0182  \\
\hline
KLIEP & 0.0015   \\
\hline
KDE & 0.0028    \\
\hline
DFW & 0.0109 \\
\hline
\end{tabular}
\caption{Feature space discrepancy between the weighted training data and the test data
}
\label{table:dis}
\end{table}

\section{Conclusion}
In this paper, we proposed the Shift Compensation Network (SCN) along with an end-to-end learning scheme for solving the covariate shift problem in citizen science.
We incorporated the discriminative loss and the feature space mean matching loss to learn the shift factor.
Tested on a real-world biodiversity-related citizen science project, \textit{eBird},
%task using the data from \textit{eBrid} citizen science project, 
we show how SCN significantly improves multi-species distribution modeling by learning and correcting for the data bias, and that it consistently performs better than previous models.
We also discovered the importance of fitting the shift factor gradually through the training process, which raises an interesting question for future research: How do the weights affect the performance of models learned by stochastic gradient descent?
Future directions include exploring the best way to learn and apply shift factors in deep learning models.

%\clearpage
\section*{Acknowledgments}
We are grateful for the work of Wenting Zhao, the Cornell Lab of Ornithology and thousands of \textit{eBird} participants.
This research was supported by the National Science Foundation (Grants Number 0832782,1522054, 1059284, 1356308), and ARO grant W911-NF-14-1-0498.

%We would like to thank Wenting Zhao for generating the heatmaps.
%We are grateful for the work of thousands of \textit{eBird} participants and for the assistance of the Cornell Lab of Ornithology. This research was supported by the National Science Foundation (Grants Number 0832782,1522054, 1059284, 1356308), and ARO grant W911-NF-14-1-0498.

%\clearpage
\small
\bibliography{dichen}
\bibliographystyle{aaai}

\end{document}